# Transformed K-means Clustering

Anurag Goel
IIIT Delhi
New Delhi, India
anuragg@iiitd.ac.in

Angshul Majumdar
IIIT Delhi
New Delhi, India
angshul@iiitd.ac.in

*Abstract*—**In this work we propose a clustering framework based on the paradigm of transform learning. In simple terms the representation from transform learning is used for K-means clustering; however, the problem is not solved in such a naïve piecemeal fashion. The K-means clustering loss is embedded into the transform learning framework and the joint problem is solved using the alternating direction method of multipliers. Results on document clustering show that our proposed approach improves over the state-of-the-art.**

*Index Terms*— clustering, K-means, representation learning

## I. INTRODUCTION

Transform learning is a relatively new representation learning framework [1-3]. It learns a basis (transform) such that it operates (analyses) on the data to generate the corresponding coefficients. It is also called analysis dictionary learning or analysis sparse coding by some [4]. Transform learning was originally developed for solving inverse problems like denoising [5] and reconstruction [6]. This may be the reason, transform learning is largely unknown outside the signal processing community.

Over the years there have been a handful of studies on transform learning based data analysis. For example [4] was used for unsupervised feature generation. It has been used for supervised feature generation [7] and classification [8]. Other variants liked kernelized transform learning [9] and robust transform learning [10]. Using transform learning as the building block, deeper versions have also been proposed for unsupervised [11, 12] and supervised [13] scenarios. Transform learning based domain adaptation formulations have also been proposed for shallow [14] and deep [15] version.

In a recent study the transform learning formulation was used for clustering [16]; therein the subspace clustering [17] was incorporated into the transform learning framework. A deeper version of the same was proposed in [18]. In both cases, the generated coefficients from the shallow or the deep versions were used as inputs for subspace clustering. In both cases [16, 18] a joint loss that incorporated both transform learning and subspace clustering was proposed and the complete formulation was solved using the alternating direction method of multipliers.

In this work, we propose to embed K-means clustering into the transform learning framework. The basic idea remains the same as in [16]. The learnt representation from transform learning is input for K-means clustering. The K-means embedded transform learning is solved as a single optimization problem. The motivation for our work arises from two observations. First, K-means is probably the most widely used general purpose clustering formulation. Second, it is a fundamental step in other clustering techniques like subspace and spectral clustering.

The rest of the paper will be organized into several sections. A brief literature review on representation learning based clustering and transform learning will be discussed in section 2. The proposed formulation is given in section 3. The experimental results will be shown in section 4. The conclusions of the work will be discussed in section 5.

## II. LITERATURE REVIEW

### A. Transform Learning

Transform learning analyses the data by learning a transform / basis to produce coefficients. Mathematically this is expressed as,

$$TX = Z \qquad (1)$$

Here $T$ is the transform, $X$ is the data and $Z$ the corresponding coefficients. The following transform learning formulation was proposed in [1] –

$$\min_{T,Z} \|TX - Z\|_F^2 + \lambda \left( \|T\|_F^2 - \log \det T \right) + \mu \|Z\|_1 \qquad (2)$$

The parameters ($\lambda$ and $\mu$) are positive. The factor $-\log \det T$ imposes a full rank on the learned transform; this prevents the degenerate solution ($T=0$, $Z=0$). The additional penalty $\|T\|_F^2$ is to balance scale.

In [1], an alternating minimization approach was proposed to solve the transform learning problem. This is given by –

$$Z \leftarrow \min_Z \|TX - Z\|_F^2 + \mu \|Z\|_1 \qquad (3a)$$

$$T \leftarrow \min_T \|TX - Z\|_F^2 + \lambda \left( \|T\|_F^2 - \log \det T \right) \qquad (3b)$$

Updating the coefficients (3a) is straightforward using one step of soft thresholding,

$$Z \leftarrow signum(TX) \cdot \max\left(0, abs(TX) - \mu\right) \quad (4)$$

Here '·' indicates element-wise product.

The update for the transform (3b) also has a closed form solution. This is given as –

$$XX^T + \lambda I = LL^T \quad (5a)$$
$$L^{-1}XZ^T = USV^T \quad (5b)$$
$$T = 0.5U\left(S + (S^2 + 2\lambda I)^{1/2}\right)V^T L^{-1} \quad (5c)$$

The proof for convergence of such an alternating update algorithm can be found in [2].

*B. Representation Learning for Clustering*

Transform learning is a relatively new framework; to the best of our knowledge, the only clustering formulation based on this framework is [16]. There the subspace clustering loss is added to the transform learning formulation (1). Its deeper version was proposed in [18].

Since dictionary learning is considered to be the synthesis version of analysis transform learning, we will briefly discuss it as well. The topic is related to the k-means type clustering [19, 20]. Dictionary learning has also been used within the subspace clustering framework [21].

Strictly speaking, dictionary learning is a specific form of matrix factorization. The equivalence of matrix factorization and spectral clustering is well known [22]; furthermore, the relationships between various matrix factorization techniques on the topic of clustering has also been studied. In [23] the factorization based formulation for K-means clustering has also been shown.

While dictionary and transform learning are shallow representation learning techniques, in current times deep representation learning is more popular in both signal processing and machine learning communities.

One of the first studies in deep learning based clustering is [24]; in there stacked autoencoder is learnt and the representation from the deepest layer is fed into a separate clustering algorithm like k-means or spectral clustering. A later study [25], embedded the (sparse subspace) clustering algorithm into the stacked autoencoder formulation. It was found that the jointly learnt formulation [25] yielded better results than the piecemeal technique [24].

Other studies like [26 - 28] were also based on ideas similar to [25]; unlike the latter which embedded sparse subspace clustering into the stacked autoencoder formulation, [26-28] incorporated the K-means algorithm. The difference between [26] and [27, 28] lies in the definition of the distance metric used in the K-means.

While all the prior studies [24-27] were based on the autoencoder formulation, [29] proposed a convolutional autoencoder based clustering technique. As in [26] Student's t-distribution based K-means clustering loss was embedded in the deepest layer of the convolutional autoencoder for segmenting the samples.

The initial work on deep clustering [24] proposed a piecemeal solution back in 2014. Over the years [25-29] it was found that jointly learnt solutions that incorporate the clustering loss into the network always improved the results. Following this observation, [30] embedded spectral clustering loss into the autoencoder based formulation.

An interesting deep clustering formulation was put forth in [31]. It proposed deep matrix factorization. The relationship between matrix factorization and clustering is well known [22-24]. The aforesaid study [31] leveraged this relationship and argued that different layers corresponded to different 'concepts' in the data. For example, if the task was face clustering, the first layer probably corresponded to clustering gender, the second layer to age, the third layer to ethnicity, and so on.

We have reviewed the major studies in deep learning based clustering. Given that it is a concise paper, we may have omitted papers that are application specific.

III. PROPOSED APPROACH

The popular way to express K-means clustering is via the following formulation:

$$\sum_{i=1}^{k}\sum_{j=1}^{n} h_{ij} \left\| z_j - \mu_i \right\|_2^2$$
$$h_{ij} = \begin{cases} 1 & \text{if } x_j \in \text{Cluster i} \\ 0 & \text{otherwise} \end{cases} \quad (6)$$

where $z_j$ denotes the $j_{th}$ sample and $\mu_i$ the $i^{th}$ cluster.

In [23], it was shown that (6) can be alternately represented in the form of matrix factorization.

$$\left\| Z - ZH^T \left(HH^T\right)^{-1} H \right\|_F^2 \quad (7)$$

where $Z$ is the data matrix formed by stacking $z_j$'s as columns and $H$ is the matrix of binary indicator variables $h_{ij}$.

In our formulation of transformed K-means clustering, the general idea is to use the coefficients generated by transform learning as inputs to K-means clustering. This is achieved by incorporating the K-means cost into the transform learning formulation.

$$\min_{T,Z,H} \underbrace{\left\| TX - Z \right\|_F^2 + \lambda\left(\left\| T \right\|_F^2 - \log \det T\right)}_{\text{Transform Learning}}$$
$$+ \mu \underbrace{\left\| Z - ZH^T \left(HH^T\right)^{-1} H \right\|_F^2}_{\text{K-means}} \quad (8)$$
$$h_{ij} = \begin{cases} 1 & \text{if } x_j \in \text{Cluster i} \\ 0 & \text{otherwise} \end{cases}$$

In (8) the transform learning formulation is regularized by the K-means cost in a single joint optimization problem. Note that, compared to the original formulation of transform learning (2), we have dropped the sparsity promoting term $Z$. This is because transform learning was originally intended to solve inverse problems so sparsity on the coefficients was necessary [32]. However, for our purpose, the $l_1$-norm on the coefficients do not carry any particular meaning apart from a regularization term. Hence we have dropped it. Note that the same has been done in other transform learning based formulations for machine learning [8-16].

The solution to (8) can be achieved via the alternating direction method of multipliers (ADMM) [33], i.e. in every iteration each of the variables is updated by assuming the others to be constant. This leads to the following three sub-problems –

$$T \leftarrow \min_T \|TX - Z\|_F^2 + \lambda \left(\|T\|_F^2 - \log \det T\right)$$

$$Z \leftarrow \min_Z \|TX - Z\|_F^2 + \mu \left\| Z - ZH^T \left(HH^T\right)^{-1} H \right\|_F^2$$

$$H \leftarrow \min_H \left\| Z - ZH^T \left(HH^T\right)^{-1} H \right\|_F^2$$

$$h_{ij} = \begin{cases} 1 & \text{if } x_j \in \text{Cluster i} \\ 0 & \text{otherwise} \end{cases}$$

The update for $T$ is the same as that of (3b); hence we do not repeat it. The solution to Z is also straightforward as one notices that it can be simplified to -

$$Z \leftarrow \min_Z \|TX - Z\|_F^2 + \mu \|ZK\|_F^2 \, ; K = I - H^T \left(HH^T\right)^{-1} H$$

Taking the derivative and equating it to 0,

$$\nabla_Z \left( \|TX - Z\|_F^2 + \mu \|ZK\|_F^2 \right) = 0$$

$$TX = Z(I + \mu K)$$

$$\Rightarrow Z = TX(I + \mu K)^{-1}$$

This concludes the closed form update for $Z$.

The solution for $H$ is straightforward; it can be obtained by K-means clustering.

K-means algorithm – Update for H

---
Initialize: Given K clusters, choose the centroids randomly from columns $Z_i^\downarrow$.

Until convergence repeat:

   Compute distance of $Z_i^\downarrow$ from each cluster centroid $\mu_j$.

   Assign $Z_i^\downarrow$ to the nearest cluster (centroid); $H_{ij} = 1$ for the said cluster and 0 otherwise.

End iterations when cluster centroids do not change (within some limits).

---

This concludes the derivation of our proposed transformed K-means algorithm. The complexity of updating T is dependent on the Cholesky and singular value decompositions; both of which have a complexity of $O(n^3)$. The update for Z has a complexity of $O(n^2)$. The K-means is ideally an NP hard problem, but the algorithm used here has a complexity of $O(t*k*n^2)$ where t is the number of loops and k the number of clusters.

### IV. EXPERIMENTAL EVALUATION

Our experiments focus on document clustering. For this application, we follow the protocol defined in a recent work [34]. For our experiments, we use three data sets TDT2 corpus [35], Reuters-21578 corpus [35], and 20 Newsgroup [35].

The TDT2 English document data set includes six months of material drawn on a daily basis from six English language news sources. In this set, the total number of samples is 9394, the feature dimension is 36771, and the number of clusters is 30.

The Reuters-21578 document set is a collection of manually categorized newswire stories from Reuters Ltd. In this set, the total number of samples is 8293, the feature dimension is 18933, and the number of clusters is 65.

The 20 Newsgroups data set is a collection of approximately 20,000 newsgroup documents. In this set, the total number of samples is 18846, the feature dimension is 26214, and the number of clusters is 20.

Following [34], we report the result in terms of two metrics namely entropy and purity. For good clustering, one requires small entropy and high purity [36]. In [34], the metrics are reported by varying the number of clusters from 2 to 10. We follow the same protocol.

Purity is given by,

$$purity = \frac{1}{n} \sum_{k=1}^{r} \max_{1 \leq l \leq q} n_k^l$$

where $n_k^l$ is the number of samples in cluster $k$ that belong to original class $l$. A larger purity value indicates better clustering performance.

Entropy measures how classes are distributed on various clusters.

$$entropy = \frac{1}{n \log_2 q} \sum_{k=1}^{r} \sum_{l=1}^{q} n_k^l \log_2 \frac{n_k^l}{n_k}$$

where $n_k = \sum_l n_k^l$. Generally, a smaller entropy value corresponds to a better clustering quality

We compare our technique with several latest approaches in document clustering. The first one is improved spherical K-means (ISKM) [37], deep embedding clustering based on contractive autoencoder (DECCA) [38] and transformed subspace clustering (TSC) [16]. Note that the deep version of TSC [18] was not used for document clustering and hence we do not consider it here. For all the datasets, across all configurations, one can see that the proposed approach yields the best results.

TABLE I. COMPARISON SHOWING IMPROVEMENT OF PROPOSED METHOD OVER EXISTING TECHNIQUES ON TDT2

| Clusters | Entropy (lower is better) | | | | Purity (higher is better) | | | |
|---|---|---|---|---|---|---|---|---|
| | ISKM | DECCA | TSC | Proposed | ISKM | DECCA | TSC | Proposed |
| 2 | **.0000** | **.0000** | **.0000** | **.0000** | **1.0000** | **1.0000** | **1.0000** | **1.0000** |
| 4 | .0059 | **.0000** | **.0000** | **.0000** | .9956 | **1.0000** | **1.0000** | **1.0000** |
| 6 | .0826 | .0809 | .0013 | **.0011** | .9435 | .9954 | .9963 | **.9971** |
| 8 | .0952 | .0911 | .0465 | **.0206** | .9476 | .9801 | .9013 | **.9901** |
| 10 | .0808 | .0685 | .0178 | **.0061** | .9153 | .9224 | .9775 | **.9854** |

TABLE II. COMPARISON SHOWING IMPROVEMENT OF PROPOSED METHOD OVER EXISTING TECHNIQUES ON REUTERS

| Clusters | Entropy (lower is better) | | | | Purity (higher is better) | | | |
|---|---|---|---|---|---|---|---|---|
| | ISKM | DECCA | TSC | Proposed | ISKM | DECCA | TSC | Proposed |
| 2 | .0551 | .0454 | .0493 | **.0451** | .9012 | .9218 | .9735 | **.9912** |
| 4 | .2751 | .2400 | .2103 | **.2008** | .8834 | .8935 | .8984 | **.9061** |
| 6 | .2029 | .2021 | .1905 | **.1435** | .8719 | .8880 | .8855 | **.8995** |
| 8 | .2158 | .2029 | .2811 | **.2009** | .8686 | .8776 | .9135 | **.9524** |
| 10 | .2677 | .2464 | .2579 | **.2286** | .7690 | .7888 | .8069 | **.8331** |

TABLE III. COMPARISON SHOWING IMPROVEMENT OF PROPOSED METHOD OVER EXISTING TECHNIQUES ON NEWSGROUP

| Clusters | Entropy (lower is better) | | | | Purity (higher is better) | | | |
|---|---|---|---|---|---|---|---|---|
| | ISKM | DECCA | TSC | Proposed | ISKM | DECCA | TSC | Proposed |
| 2 | .1556 | .1843 | .8172 | **.1131** | .8867 | .8200 | .7233 | **.9233** |
| 4 | .1665 | .1301 | .5911 | **.1055** | .8083 | .8183 | .6567 | **.8575** |
| 6 | .1441 | .1492 | .4697 | **.1211** | .8717 | .8322 | .7050 | **.9028** |
| 8 | .1505 | .1562 | .4673 | **.1142** | .8708 | .8700 | .6721 | **.8859** |
| 10 | .1395 | .1625 | .4449 | **.1299** | .8943 | .8677 | .6690 | **.9100** |

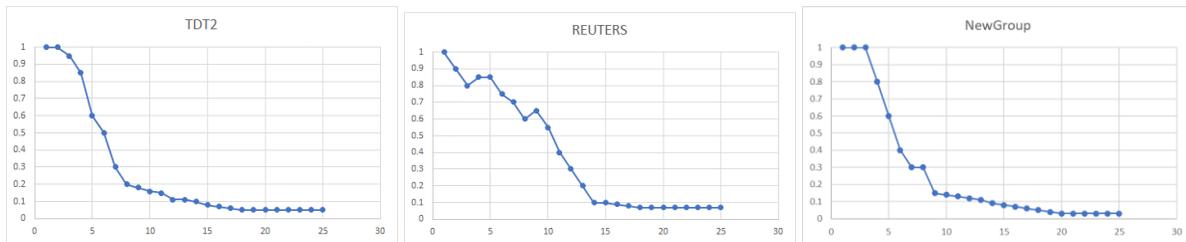

**Fig. 1.** Empirical Convergence Plot (2 clusters) for Proposed Algorithm

The empirical convergence plot of the proposed approach is given in Fig. 1. The results are shown for two clusters. One can observe that for REUTERS the convergence is not monotonic. This can be the case for ADMM based solutions. We see that our algorithm converges within 15 to 20 iterations. This has been the case irrespective of the number of clusters.

## V. CONCLUSION

This work proposes the formulation of transformed K-means clustering; it embeds K-means clustering into the transform learning framework. The ensuing formulation is solved via ADMM. Results on document clustering show promise; our method improves over the state-of-the-art in this area.

In the future, we would like to extend this work and embed K-means clustering into the deep transform learning framework [12]. As has been seen for the case of transformed subspace clustering [16], its deeper version [18] improves over the shallower one. For K-means clustering we can similar results by going deeper.

Another area where we can extend this work is by kernelizing the transform [10]. In the past [16], non-linear kernels improved clustering results by a significant margin.